\documentclass[letterpaper]{article} 
\usepackage{aaai23}  
\usepackage{times}  
\usepackage{helvet}  
\usepackage{courier}  
\usepackage[hyphens]{url}  
\usepackage{graphicx} 
\usepackage{natbib}  
\usepackage{caption} 
\usepackage{algorithm}
\usepackage{algorithmic}

\usepackage{newfloat}
\usepackage{listings}

\usepackage{subcaption}
\usepackage{paralist}

\DeclareCaptionStyle{ruled}{labelfont=normalfont,labelsep=colon,strut=off} 
\floatstyle{ruled}
\newfloat{listing}{tb}{lst}{}
\floatname{listing}{Listing}
\title{Fairly Private: Investigating The Fairness of Visual Privacy Preservation Algorithms}
\author {
	Sophie Noiret,\textsuperscript{\rm 1}
	Siddharth Ravi, \textsuperscript{\rm 2}
	Martin Kampel, \textsuperscript{\rm 1}
	Francisco Florez-Revuelta \textsuperscript{\rm 2}
}
\affiliations {
	\textsuperscript{\rm 1} Vienna University of Technology \\
	\textsuperscript{\rm 2} University of Alicante\\
	snoiret@cvl.tuwien.ac.at.com, siddharth.ravi@gcloud.ua.es, martin.kampel@tuwien.ac.at, francisco.florez@gcloud.ua.es
}

\begin{document}
	
	\maketitle
	
	\begin{abstract}
		As the privacy risks posed by camera surveillance and facial recognition have grown, so has the research into privacy preservation algorithms. Among these, visual privacy preservation algorithms attempt to impart bodily privacy to subjects in visuals by obfuscating privacy-sensitive areas. While disparate performances of facial recognition systems across phenotypes are the subject of much study, its counterpart, privacy preservation, is not commonly analysed from a fairness perspective. In this paper, the fairness of commonly used visual privacy preservation algorithms is investigated through the performances of facial recognition models on obfuscated images. Experiments on the PubFig dataset clearly show that the privacy protection provided is unequal across groups.
	\end{abstract}
	
	\section{Introduction}

	According to the website IFSEC Global\footnote{https://www.ifsecglobal.com/video-surveillance/whats-behind-the-global-growth-of-cloud-based-video-surveillance/, last accessed September 26, 2022}, the worldwide video surveillance camera market is expected to almost double by 2025 compared to 2019. It is therefore of little surprise that video surveillance is being considered worldwide as a serious threat to privacy \cite{SensorsAndSensibility}. Surveillance is, however, not always malicious, and could instead be a necessity. In medical settings, for instance, the capacity to monitor patients remotely can be crucial to ensuring their safety. In these cases, while knowing that a person is present in the image is useful, it might not be necessary to know the identity of the person in the image. Visual privacy preservation algorithms can therefore be deployed to protect bodily privacy in such settings. Privacy preservation algorithms (PPA) work by perceptually obfuscating the visual feed to various degrees depending on the context. Some of the simplest and most commonly used visual PPAs include blurring and pixelation. 
	While these algorithms have been in use for decades, there have been various issues surrounding their use. Simple algorithms are also prone to attacks with which the obfuscation performed can be reversed, and the original image reconstructed~\cite{korshunov2011video,Menon_2020_CVPR}. Facial recognition algorithms can also be trained to identify subjects present in obfuscated images. Furthermore, while the biases in facial recognition are well documented~\cite{pmlr-v81-buolamwini18a}, the effectiveness of obfuscation has not been subjected to the same amount of scrutiny from a fairness perspective. 

	The primary questions explored in this work are:
	\begin{compactitem}
		\item Is there a racial or gender bias in the degree of privacy afforded by face obfuscation techniques? 
		\item If so, does this bias depend on the obfuscation method used? The classifier used for facial recognition? 
	\end{compactitem}
	These questions are examined under the assumption that people do not wish to be recognized, i.e., that a positive result is one in which the subject is not correctly identified. 
	In this work, a system is deemed fair if it achieves equal performance across groups, and one's privacy is considered to be protected if their identity is successfully concealed by a face obfuscation method. 
	
	This work studies discrimination based on the protected attributes of race and gender. By training a facial recognition system on un-obfuscated images and setting it to predict on obfuscated images, the capacity of blurring and pixelation to fairly conceal identities is evaluated. The results show that both of these techniques yield significantly disparate performances across demographics. To quantify experimental results better, this paper also introduces a new metric called the \textit{personal detection rate}.
	
	The rest of this paper is structured as follows. Section \ref{sec:related} examines related work in the areas of fairness and privacy in computer vision.  
	Section \ref{sec:methodology} introduces and motivates the methodology used for choosing the dataset selected, and the framework used for analysis in this paper. 
	This section also presents the experiments conducted, as well as details about their implementation. Section \ref{sec:results} explains the results obtained from these experiments and discusses factors of importance that influence the results. Finally, in Section \ref{sec:conclusion} the paper puts forth its conclusions, discusses the limitations of the experiments, and proposes various possible avenues for future research.
	
	\section{Related Work}
	\label{sec:related}
	Several works have analysed the intersection of fairness and algorithmic privacy \cite{ekstrand2018privacy,dwork2013s,10.1145/3314183.3323847}. The existing literature at this intersection relevant to this paper can be separated into two categories. These are works that primarily deal with the issue of fair privacy, and those dealing with the topic of fairness in facial analysis.
	\subsection{Fair Privacy}
	Prior works have studied the intersection of fairness and privacy by using sensitive attributes that are not of a visual nature \cite{Ding_Zhang_Li_Wang_Yu_Pan_2020,Tran_Fioretto_Van_Hentenryck_2021,Ghili_Kazemi_Karbasi_2019}. Research, for example, has been conducted on the impact of fairness in the area of privacy protected data (through differential privacy) in the domains of voting rights and funds allocation \cite{pujol2020fair}. This highlights the need to consider the fairness of outcomes when designing privacy algorithms.
	
	Models that provide both privacy and fairness from a more theoretical standpoint have also been explored. For example, the creation of two logistic regression models (PFLR and PFLR*) that are differentially private and provide fairness have been explored, which at the same time preserves the utility of the resulting model \cite{xu2019achieving}.
	
	\subsection{Fair Facial Analysis}
	The fairness provided by software performing facial analysis has come under scrutiny\cite{VendorTest}. However, works such as \cite{RFW} and \cite{Latent} aim at improving the fairness of facial recognition technology, while this paper analyses a scenario in which subjects do not wish to be identified. 
	In their work 'Gender Shades', Buolamwini et al.~\cite{pmlr-v81-buolamwini18a} evaluate 3 commercial gender classification systems and show that darker skinned females are the most misclassified group. The authors also analyse two facial analysis benchmark datasets, IJB-A \cite{Klare_2015_CVPR} and Adience \cite{6906255}, and find that the composition of these datasets is overwhelmingly made up by light-skinned individuals. 
	
	Methods in the literature aiming to achieve fairness in facial recognition work by creating fairer facial embeddings. Alvi et al. \cite{Alvi_2018_ECCV_Workshops}, for example, created a facial recognition method that calculates the cross entropy between the output distributions produced by classifiers trained on biased data and a uniform distribution. This is then shown to be a fairer feature representation for the task of facial recognition. 
	
	Neural networks which work by imparting fairness at the comparison level have also been created for the task of facial recognition \cite{ComparisonLevel}. This is achieved by learning a similarity function model which treats people from different ethnicities similarly. This produces fair comparison scores when presented with biased face embeddings. For this, a neural network is trained with a loss function which includes a penalization term prioritizing group fairness or individual fairness.
	Although the paper puts forth a method to improve fairness in facial recognition, it does not provide results for privacy-protected images. It also assumes a desire for fairness on the part of the recognition system, which we do not.

	Adversarially trained models that discourage facial recognition networks from encoding information about protected attributes have also been explored~\cite{Dhar_2021_ICCV}. This work, however, also is not tested on privacy-protected facial images, but rather uses unobfuscated image data to validate the study.
	
	Masked facial recognition is a field that is of interest to the topic of this paper. Yu et al.~\cite{boostingFairnessMasked} create a method to improve the fairness of masked face recognition algorithms. 
	Unlike our work, however, the context for analysis is one in which higher recognition rates are the desired outcome, and the faces are only partially obfuscated. %
	
	This paper, in contrast to these prior works, seeks specifically to study the fairness of commonly used visual privacy preservation algorithms. To the best of the authors' knowledge, such a study has not been attempted. 
	
	\section{Methodology and Experiments}
	\label{sec:methodology}
	The scenario motivating this paper is one in which a surveillance camera is placed in a building. This specific use-case necessitates to be able to know whether a face is in the picture or not, but knowing the identity of the person is not required. Consequently, people's privacy is protected by obfuscating the face of the person in the image. A bad actor with access to unobfuscated images of the people in the building (for instance from employee files) can then train a model to recognize the faces in the obfuscated images. It is also impossible to imagine the use of adversarial noise to thwart machine learning models in a scenario such as this, mainly due to the lack of computational power in a typical setup. 
	\subsection{Dataset}
	
	\begin{figure}[t]
		\centering
		\begin{subfigure}{0.23\textwidth}
			\includegraphics[width=\columnwidth]{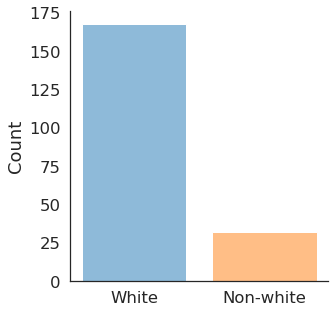}
			\caption{\small By race (binary)}
			\label{Obf-a}
		\end{subfigure}
		\hfill
		\begin{subfigure}{0.45\columnwidth}
			\includegraphics[width=\linewidth]{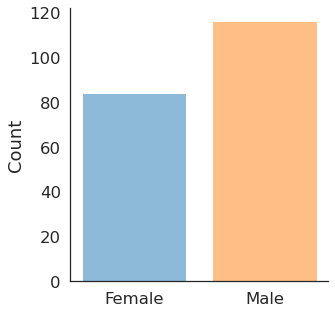}
			\caption{\small By gender}
			\label{Obf-c}
		\end{subfigure}
		\hfill
		\begin{subfigure}{0.45\columnwidth}
			\includegraphics[width=\linewidth]{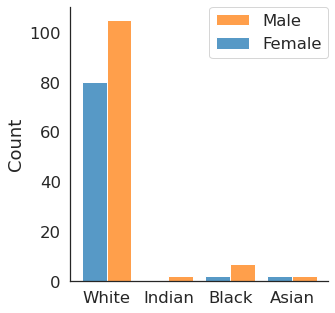}
			\caption{\small By race and gender}
			\label{ObfRace-d}
		\end{subfigure}
		\hfill
		\begin{subfigure}{0.45\columnwidth}
			\includegraphics[width=\linewidth]{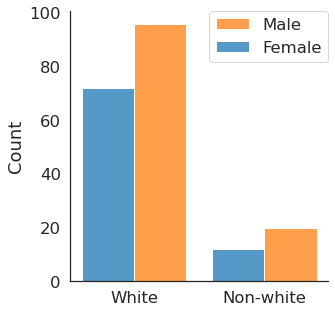}
			\caption{\small By race (binary), gender}
			\label{ObfGender-e}
		\end{subfigure}
		\caption{\small Composition of Dataset}
		\label{fig:ObfuscationBlur}
	\end{figure}
	\begin{figure*}[t]
		\centering
		\includegraphics[width=\linewidth]{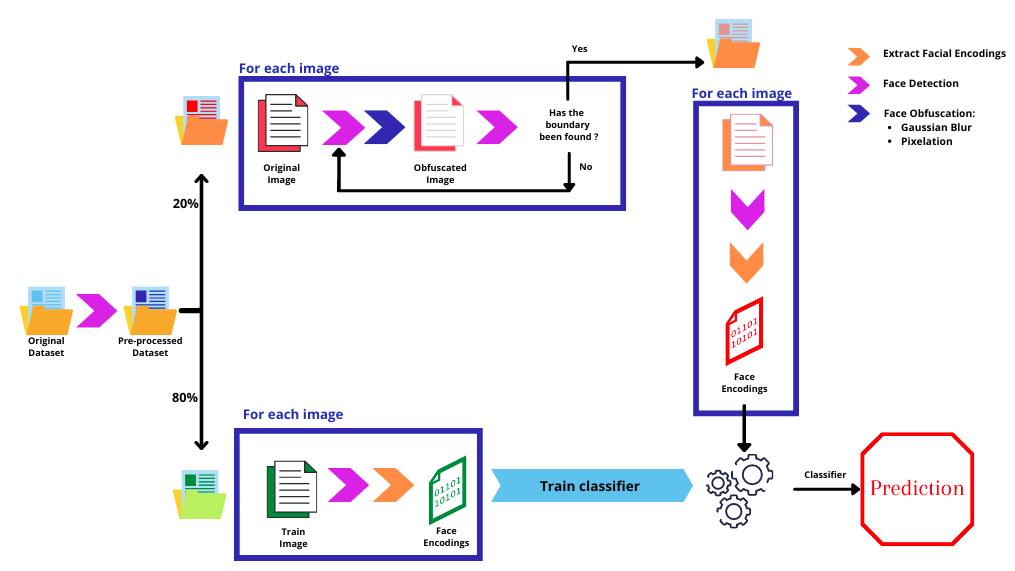}
		\caption{\small Method overview}
		\label{Pipeline}
	\end{figure*}
	
	The requirements to conduct this work dictates that the dataset used contains images annotated for identity and protected characteristics, as well as containing several images per subject. Having a balanced dataset with race and gender information is not essential, as nothing guarantees such a balance in our scenario. 
	Based on these criteria, the PubFig dataset \cite{5459250} is chosen. This dataset contains 58,797 images of 200 people according to the original composition and is available as a list of URLs due to copyright issues. As the dataset is from 2008, many of the original URLs are broken and the corresponding images are consequently excluded. Attribute labels are also provided along with the dataset to facilitate research, containing various protected attributes such as racial categories and gender. According to the dataset documentation provided, the attribute labels have been partially acquired through Amazon Mechanical Turk workers and partially generated by an attribute classifier. As a consequence, the dataset contains several errors in the attribute labels provided along with the dataset. This dataset also contains the additional difficulty of varying picture quality. For this reason, we do not apply an uniform level of obfuscation. The details are provided in the next section.  
	Although other datasets are used in face recognition experiments, but they either lack the necessary demographic annotations (CelebA \cite{CelebA}, VGGFace \cite{VGGFace}) or identity information (FairFace \cite{Karkkainen_2021_WACV}).
	To create the attribute labels required for the experiments, the dataset has been cleaned and aggregated to obtain per-individual data. The original composition of the dataset can be seen in Fig. \ref{fig:ObfuscationBlur}. Due to the labelling errors in the original set, protected attributes have been manually checked and corrected if necessary. As can be observed, the original composition is severely imbalanced, with a large majority of the people in the set being white. While this does not prohibit experimentation, it leads to some groups (for instance, Indian women) not having sufficient representation for analysis. For this reason, the final attribute labels are chosen to be binary, i.e., race (white vs non-white) and gender (male vs female).The race groups chosen as reference to decide white vs non-white individuals are as defined in the FairFace dataset \cite{Karkkainen_2021_WACV}.

	\subsection{Process overview}
	
	The main steps of the experiment are executed according to the following pipeline:
	\begin{compactenum}
		\item From the original dataset, faces are detected on each image to only keep images with exactly one face.
		\item For each person in the dataset, a random 80/20 split is performed. 
		\item 20\% of images are obfuscated. On these images, face detection is performed and a 128-dimensional vector of face encodings is created for each image. 
		\item On the remaining 80\%, face detection is performed and a 128-dimensional vector of face encodings is created for each image.
		\item A classifier is trained on the encodings of the unobfuscated images.
		\item This classifier is subsequently used to predict the identity of the person present in the obfuscated images. 
	\end{compactenum}
	Considering the varying quality of the pictures present in the dataset, a different level of obfuscation is required per image on step 3. For the case analysed in this paper, it is necessary to know that a face is in the picture. Consequently, we choose the maximum level of obfuscation that still allows for face detection. This means that potential biases in face detection will result in a lower level of obfuscation: the complete system (face detection and face obfuscation) will therefore reflect bias in both tasks. To determine this level of obfuscation, each image goes through the following process: 
	\begin{compactitem}
		\item A face is detected in the image and obfuscated,  and face detection is run again on the image. 
		\item If the face is not detected, then while no face is detected, the obfuscation level is cut by half and a new obfuscation takes place. 
		\item If a face is  detected, then while a face is detected, the obfuscation level is doubled and a new obfuscation is performed.
	\end{compactitem}
	This gives us a range, providing a maximum (when the face is not detected) and a minimum (when the face is detected) level of obfuscation. A binary search is then performed on this interval to find the upper limit of obfuscation that still allows for face detection in each image. This process can be seen in Fig. \ref{Pipeline}.

	\subsection{Experiments Conducted}
	\label{sec:experiments}
	The main steps of the pipeline are implemented through several techniques to ascertain their influence on any potential bias that would emerge.
	
	\textbf{Pre-processing of dataset.} 
	The original dataset includes a checksum computed using the md5sum command for each image.
	In an effort to avoid having the same picture repeated in the set, only one image per checksum is kept. To obtain per-image results, face detection is performed on every image in the dataset and only images with exactly one face are retained for the experiments.
	
	\textbf{Face Detection.}
	For face detection, the method used is based on the Histogram of Oriented Gradients (HOG) as implemented in the dlib and face\_recognition\footnote{Available at \url{https://github.com/ageitgey/face_recognition}} libraries. HOG is a feature descriptor trained on the Labeled Faces in Wild \cite{huang:inria-00321923} dataset to detect human faces.

	\begin{figure}[h]
		\small
		\centering
		\begin{subfigure}{0.3\columnwidth}
			\centering
			\includegraphics[width=\linewidth]{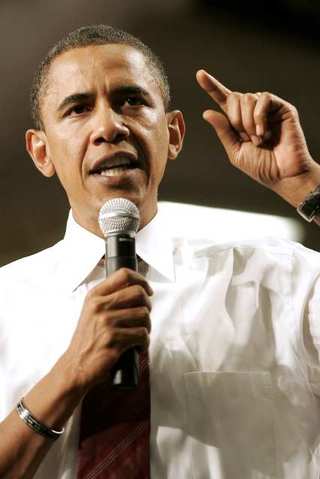}
			\caption{\small Original}
			\label{OrigImage}
		\end{subfigure}
		
		\begin{subfigure}{0.3\columnwidth}
			\centering
			\includegraphics[width=\linewidth]{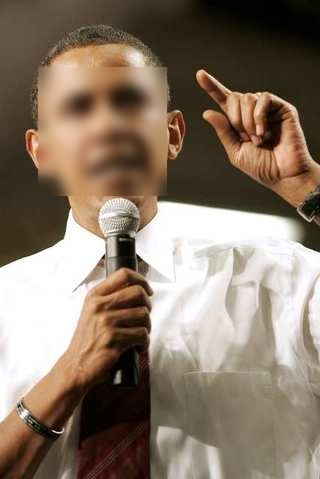}
			\caption{\small Blur}
			\label{BlurredHOGImage}
		\end{subfigure}
		\begin{subfigure}{0.3\columnwidth}
			\centering
			\includegraphics[width=\linewidth]{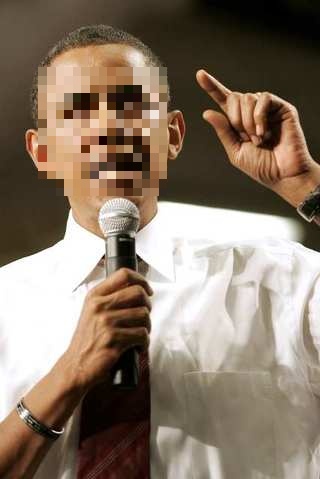}
			\caption{\small Pixelated}
			\label{PixelImage}
		\end{subfigure}
		\caption{\small Original and modified images} 
		\label{ObfMethods}
	\end{figure}
	
	\textbf{Face obfuscation.}The two face obfuscation techniques used are Gaussian blurring and pixelation. Gaussian blurring masks details by considering for each pixel the value of the pixels in its neighbourhood. The section around each pixel, called the kernel, is used to modify its value. A bigger kernel leads to more blurring. Gaussian blur, implemented in the OpenCV function \textit{GaussianBlur}, is used for the experiments in this paper. \\
	Pixelation consists of downsizing the image, then re-sizing it to the original size. 
	When downsizing the image to a smaller size, pixels are deleted. These pixels are replaced when the image is restored to its original size by interpolating new pixels, which are calculated according to an interpolation method chosen. 
	Pixelation is implemented for the experiments in this paper through the \textit{resize} function provided in the OpenCV library.    
	These techniques are illustrated in Fig. \ref{ObfMethods}.
	
	\textbf{Classifier.}
	The four classifiers used to make predictions in this study are the K-Nearest Neighbour (KNN), the Naïve Bayes (NB), the Support Vector Classifier (SVC) and the Multi-Layer Perceptron (MLP). These classifiers are selected because they are commonly used multi-class classifiers. The implementation used is as in the \textit{Scikit-learn} library, with parameters either left at default (NB, SVM), or chosen through the execution of a grid search.
	
	\textbf{Measuring Algorithmic Fairness.} The fairness definition chosen here is group fairness, i.e., equal or unequal performance across groups. Performance in this case is the capacity of the privacy-preserving method to obfuscate someone's identity, which means that the favourable outcome is one where the person is not being recognized. Consequently, contrary to previous works, low accuracy, precision, recall and F1-score are the desired outcomes. These metrics are all reported for completeness; however, they can lead to different conclusions as they place emphasis on different aspects. When considering both race and gender, we also report bias, i.e., the biggest gap in performance between any two groups for each metrics.
	We make the assumption that people prefer not to be identified, neither correctly nor incorrectly. As a consequence, it is desirable to have low numbers of both the True Positives (TP, implying a correct identification), and False Positives (FP, implying an incorrect identification of the person). As accuracy reports the proportion of correct predictions, both TP and True Negatives (TN) increase the accuracy. Additionally, it is sensitive to class imbalance. 
	Therefore, when accuracy leads to a different conclusion than the other metrics, precision and recall are given preference. 
	We also report a new metric called \textit{Personal Detection Rate} (PDR), given by the following formula -
	
	\begin{equation}
	\textnormal{PDR} = \frac{\textnormal{True Positives} + \textnormal{False Positives}}{\textnormal{Number of predictions}} 
	\end{equation}
	
	This metric corresponds to  the number of times the person is identified in a picture (correctly or incorrectly), divided by the number of predictions made. However, it is also sensitive to class imbalances. 
	
	\section{Results and Discussion}
	\label{sec:results}

	The results presented in the following sections are obtained by using SVC. For reference, we consider that the groups on which the worst results are obtained are those that score higher on recognition metrics. Best results are shown in bold, while the worst results are in italics.
	
	\subsection{Level of Obfuscation Needed}
	
	\begin{figure}[h]
		\centering
		\begin{subfigure}{0.3\textwidth}
			\includegraphics[width=\linewidth]{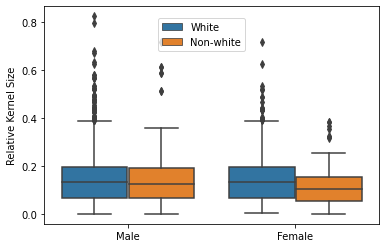}
			\caption{\small Blurring by race and gender}
			\label{Obf}
		\end{subfigure}
		\hfill
		\begin{subfigure}{0.3\textwidth}
			\includegraphics[width=\linewidth]{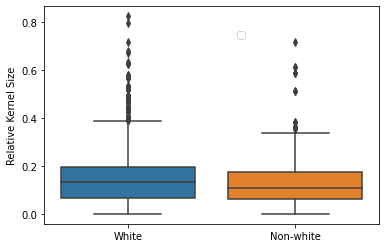}
			\caption{\small Blurring by race}
			\label{ObfRace}
		\end{subfigure}
		\hfill
		\begin{subfigure}{0.3\textwidth}
			\includegraphics[width=\linewidth]{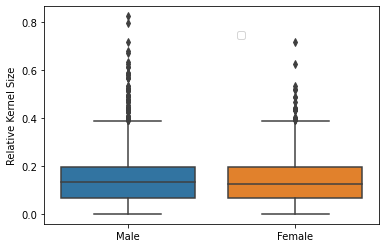}
			\caption{\small Blurring by gender}
			\label{ObfGender}
		\end{subfigure}
		\caption{\small Distribution of the level of Blurring}
		\label{ObfuscationBlur}
	\end{figure}
	
	\textbf{Gaussian Blur.}
	The amount of obfuscation performed by Gaussian blur is based on the size of the kernel. The size of the kernel is proportional to the blurring performed on the image. This value is in pixels, so it is normalized by dividing it by the size of the Region Of Interest (ROI), the face, in pixels. This aims at reducing the influence of the quality of the original picture.\\ 
	We also observe that for non-white people, the level of obfuscation is lower than it is for white people (median value of 0.109 and 0.133 respectively). This might be because the dlib library is utilized for its HOG implementation, which is pre-trained on the LFW dataset. This dataset is shown in \cite{Karkkainen_2021_WACV} to be imbalanced towards white faces. The level of obfuscation differs also between men and women: the median value is 0.132 on men and 0.126 on women.\\
	
\begin{figure}[h]
		\centering
		\begin{subfigure}{0.4\columnwidth}
			\includegraphics[width=0.9\linewidth]{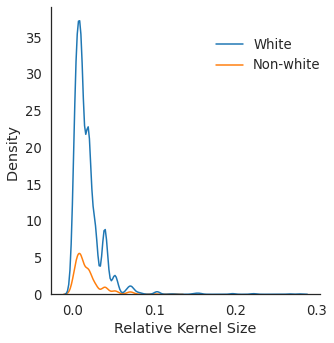}
			\caption{\small Pixelation by race}
			\label{PixelRaceZoom}
		\end{subfigure}
		\begin{subfigure}{0.4\columnwidth}
			\includegraphics[width=0.9\linewidth]{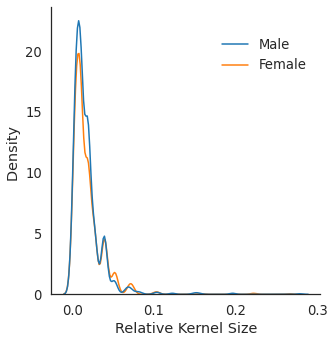}
			\caption{\small Pixelation by gender}
			\label{PixelGender}
		\end{subfigure}
		\caption{\small Distribution of the level of Pixelation}
		\label{Pixelation}
	\end{figure}

	\textbf{Pixelation.}
	Pixelation, as previously mentioned, is done by downsampling and then subsequently upsampling the region of interest. The level of obfuscation is the size to which it is downsampled: the smaller the size, the more pixelated the result is. This value is also divided by the size of the ROI to normalize it. 
	The results are presented in Fig \ref{Pixelation}. To have a clearer look at the results by race, we present a truncated version which excludes relative kernel sizes over 0.3. This excludes 16 images of white people, 12 women and 4 men. \\
	The median level of pixelation is the same between white and non-white people and between men and women, with a value of 0.0107 for all.

	\subsection {Bias in Face Recognition Results}
	The only results presented in the following sections are obtained by using SVC for the sake of brevity. For reference, we consider that the groups on which the worst results are obtained are those that score higher on recognition metrics. Best results are shown in bold, while the worst results are in italics.The row "Bias" refers to the biggest gap between two groups.

	\begin{table*}[]
		\centering
		\begin{tabular}{lrrrrr}
			
			{} &  Balanced Accuracy &  Recall &  Precision &  F1-score &  PDR \\
			\hline
			Overall  &   0.359   &   0.415   &   0.505   &   0.404 &  \\
			\hline	
			White               &              \textbf{0.357} &   0.416 &      \textbf{0.527} &     \textbf{0.414} &       0.887 \\
			Non-White           &              0.370 &   \textbf{0.407} &      0.718 &     0.475 &       \textbf{0.113} \\
			\hline	
			Male                &              0.371 &   0.439 &      0.567 &     0.438 &       0.570 \\
			Female              &              \textbf{0.342} &   \textbf{0.388} &      \textbf{0.471} &     \textbf{0.385} &       \textbf{0.430} \\
			\hline	
			Non-White Female  &              \textit{0.452} &   0.428 &      \textit{0.767} &     \textit{0.510} &       0.063 \\
			Non-White Male    &              \textbf{0.321} &   0.389 &      0.694 &     0.453 &       \textbf{0.051} \\
			White Female      &              0.323 &   \textbf{0.381} &      \textbf{0.492} &     \textbf{0.388} &       0.368 \\
			White Male        &              0.382 &   \textit{0.447} &      0.588 &     0.453 &       \textit{0.519} \\
			Bias              &             0.131  &   0.066       &       0.275       &       0.122      &       0.012  \\
			\hline
			
		\end{tabular}		
		\caption{Pixelation obfuscation, SVC classifier}
		\label{HOGPixel}
	\end{table*}
	
	\begin{table*}[]
		\centering
		\begin{tabular}{lrrrrr}
			{} &  Balanced Accuracy &  Recall &  Precision &  F1-score &  PDR \\
			\hline	Overall  &   0.333   &   0.398   &   0.556   &   0.398  &   \\
			\hline
			White               &              \textbf{0.300} &   \textbf{0.371} &      \textbf{0.579} &     \textbf{0.393} &       0.757 \\
			Non-White           &              0.508 &   0.557 &      0.793 &     0.613 &       \textbf{0.243} \\
			\hline	
			Male                &              0.366 &   0.438 &      0.628 &     0.459 &       0.531 \\
			Female              &              \textbf{0.289} &   \textbf{0.356} &      \textbf{0.516} &     \textbf{0.350} &       \textbf{0.469} \\
			\hline	
			Non-White Female  &              0.497 &   0.514 &      0.765 &     0.576 &       0.138 \\
			Non-White Male    &              \textit{0.514} &   \textit{0.592} &      \textit{0.867} &     \textit{0.669} &       \textbf{0.105} \\
			White Female      &              \textbf{0.255} &   \textbf{0.330} &      \textbf{0.543} &     \textbf{0.348} &       0.331 \\
			White Male        &              0.335 &   0.409 &      0.646 &     0.449 &       \textit{0.427} \\
			Bias              &              0.259 &   0.262 &      0.324 &     0.321 &       0.322 \\
			\hline
		\end{tabular}
		\caption{Blurring obfuscation, SVC classifier}	
		\label{HOGBlur}
	\end{table*}	
	
	The results are shown in Tables \ref{HOGPixel} and \ref{HOGBlur}.  
	
	When using pixelation for face obfuscation, the groups that get recognized at the lowest rate are women, white people, and specifically, white women. When looking at intersectional results, the group that gets the worst results are non-white women. These results are consistent across all classifiers.   
	
	When using blurring for face obfuscation, the groups that get recognized at the lowest rate are women, white people, and specifically, white women. These results are consistent across all classifiers. When looking at intersectional results, the groups that get the worst results are non-white men (SVC, MLP) and non-white women (NB, KNN).  
	
	These results show a bias in identity obfuscation for both pixelation and blurring. While women, white people, and white women get systemically better results, the group that gets worse results is not systematically non-white men. This highlights the need for intersectional studies, without which the bias against non-white women would be concealed.   
	
	\begin{figure}[h]
		\centering
		\includegraphics[width=0.9\columnwidth]{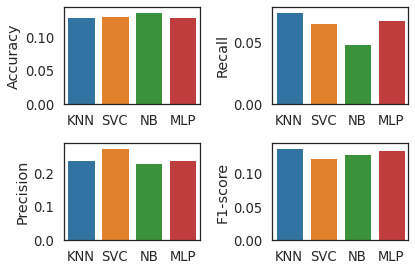}
		\caption{\small Bias across classifier, Pixelation}
		\label{biasPixel}
	\end{figure}
	
	\subsection{Influence of the classifier}
	
	As several metrics are used on different groups, it is not possible to clearly determine which classifier achieves the best performance: there is not one classifier which consistently performs better on all metrics and across all groups. Moreover, the goal here is not to determine which classifier is better, but if trends observed are consistent regardless of the classifier used.
	
	\textbf{On performance.}
	When using pixelation for face obfuscation, the difference in performance is at most 7\%. (between the precision on non-white females when using SVC and KNN).
	On all other metrics, across all different groups, the difference in performance between classifier is less than 4\%. 
	When using blurring for face obfuscation, SVC performs overall slightly better than the other classifiers. However, even here, the biggest difference in performance (difference between the precision on non-white males when using NB and SVC) is only 11\%, with the difference on all other metrics, across all different groups being less than 6\%. Overall, the influence of the classifier on performance is minimal. 
	
\begin{figure}
		\centering
		\includegraphics[width=0.9\columnwidth]{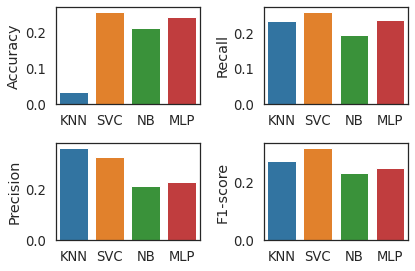}
		\caption{\small Bias across classifier, Gaussian Blur}
		\label{biasBlur}
	\end{figure}

	\textbf{On bias.}
	General trends are independent of classifier. When controlling for other factors, the groups on which the best and worst performances are achieved stay the same. 
	As can be seen in Fig. \ref{biasPixel}, the reduction of bias depends on which metric is considered when using the pixelation. However, the biggest variation in bias (between SVC and NB on precision) is only 0.045 for pixelation and while SVC performs better on recall, precision and F1-score, it performs worse on accuracy. When using blurring as shown in Fig. \ref{biasBlur}, NB is overall less biased. However, in the scenario considered, the choice of classifier is that of the bad actor, and not of the entity trying to fairly preserve people's identity. 
	\subsection{Influence of the face obfuscation methods}
	
	The face obfuscation method does not influence the general trend: the best results are obtained on white people, women and white women. The worst results are obtained on non-white people for pixelation, on non-white men (MLP and SVC) and non-white women (NB) for blurring. 
	However, as we can see in Fig \ref{biasBlur} and Fig \ref{biasPixel}, using pixelation for face obfuscation leads to a smaller gap in performance across all metrics and for all classifiers.   
	
	\section{Limitations, Conclusions, and Future Work}
	\label{sec:conclusion}
	Our work focuses intentionally on an imbalanced dataset. However, a similar study on a balanced dataset could help determine the origin of the bias shown in this work. This has not, however, been attempted because of the low number of persons in the used dataset, and also the low number of images present for some persons in the dataset. The authors note that the creation of a bigger, balanced dataset with the characteristics of PubFig would be highly beneficial to push forward research.  
	Additionally, the choice of iteratively obfuscating faces to find the limit of face detection is not one that could easily be implemented in a real-time system. 
	The reproducibility of this study is also impeded by the fact that, like Kumar et al.~\cite{5459250}, we are unable to distribute image files. However, the code developed for this study will be made available on Github.
	
	As has been shown, the degree to which the identity is protected by face detection is subject to racial and gender bias. This bias is present regardless of the classifier of face obfuscation technique used, but using pixelation instead of blurring leads to less bias.
	
	The scope of this study could be broadened in future work by performing a similar experiment on a dataset balanced with regard to race and gender, or by considering other protected attributes and face obfuscation techniques. It could also be extended to other types of person recognition, such as whole body recognition, and different types of images, such as depth data.

	\section{Acknowledgements}
	This work is part of the visuAAL project on Privacy-Aware and Acceptable Video-Based Technologies and Services for Active and Assisted Living~(\url{https://www.visuaal-itn.eu/}). This project has received funding from the European Union’s Horizon 2020 research and innovation program under the Marie Skłodowska-Curie grant agreement No 861091. It was also partly supported by the Austrian Research Promotion Agency (FFG) under the grant agreement No. 878730 and the Wiener Wissenschafts-, Forschungs- und Technologiefonds (WWTF) under the grant number ICT20-055.

	\bibliography{aaai23}

\end{document}